\documentclass[letterpaper, 10pt, conference]{IEEEtran}  

\usepackage{times}
\usepackage{graphicx}
\usepackage{epstopdf}

\usepackage{mdwmath}
\usepackage{mdwtab}
\usepackage{rotating}
\usepackage{caption}
\usepackage{subcaption}

\usepackage{amssymb}
\usepackage{amsfonts}
\usepackage{amsmath}
\usepackage{amsthm}
\usepackage{bm}

\usepackage{algorithm}
\usepackage{cite}

\usepackage{multirow} 
\usepackage{multicol}
\usepackage{array}

\usepackage{standalone}
\usepackage{booktabs} 

\usepackage{siunitx} 

\usepackage{accents}

\usepackage{pgfplotstable} 
\usepackage{verbatim}
\usepackage{isomath}
\usepackage{overpic}
\usepackage{tikz}
\usetikzlibrary{shapes,calc,patterns,
	decorations.pathmorphing,
	decorations.markings}

\tikzstyle{spring}=[very thick,decorate,decoration={zigzag,pre length=2,post
	length=2,segment length=6}]

\tikzstyle{damper}=[thick,decoration={markings, 
	mark connection node=dmp,
	mark=at position 0.5 with 
	{
		\node (dmp) [thick,inner sep=0pt,transform shape,rotate=-90,minimum
		width=15pt,minimum height=3pt,draw=none] {};
		\draw [thick] ( $(dmp.north east)+(2pt,0)$ ) -- (dmp.south east) -- (dmp.south
		west) -- ( $(dmp.north west)+(2pt,0)$ );
		\draw [thick] ( $(dmp.north)+(0,-5pt)$ ) -- ( $(dmp.north)+(0,5pt)$ );
	}
}, decorate]

\makeatletter\newcommand{\manuallabel}[2]{\def\@currentlabel{#2}\label{#1}}\makeatother

\usepackage{color}
\usepackage{xcolor}

\colorlet   {lightorange}{orange!20}
\colorlet   {lightgrey}  {gray!20}

\usepackage{amssymb}
\usepackage{mathtools}

\mathchardef\mhyphen="2D   

\newcommand{\RNum}[1]{\uppercase\expandafter{\romannumeral #1\relax}}

\graphicspath{
{figures/}
}

\providecommand{\figurename}{Fig.}

\usepackage[inline]{enumitem}
\setlist{nolistsep}
\newcommand{\il}[1]{\begin{enumerate*}[label=(\roman*)]#1\end{enumerate*}}

\newcommand*{\schemeref}[1]{Scheme \ref{scheme:#1}} 

\usepackage[normalem]{ulem}                                        
\usepackage{marginnote}
\usepackage[textwidth=10ex,colorinlistoftodos]{todonotes}

\colorlet{fwu}{red}
\colorlet{ywu}{blue}
\colorlet{zbing}{green}

\usepackage{graphics} 
\usepackage{graphicx}
\usepackage{epsfig} 
\usepackage{times} 

\usepackage{amsmath} 

\usepackage{physics}
\usepackage{xcolor}
\usepackage{tikz}
\usetikzlibrary{calc} 
\usetikzlibrary{positioning}
\usepackage{balance}
\usepackage{soul}
\usetikzlibrary{arrows}

\usepackage{algorithm}
\usepackage{algpseudocode}

\usepackage[top=2.54cm, bottom=1.91cm, left=1.91cm, right=1.91cm]{geometry}

\usepackage{multirow}
\usepackage{array}
\usepackage{wrapfig}
\usepackage{tabularx}
\usepackage[flushleft]{threeparttable}
\usepackage{subcaption}

\usepackage{float}

\usepackage{hyperref}

\newcommand{\mysection}[2]{
  \noindent \textbf{#1} #2
}

\IEEEoverridecommandlockouts

\title{\LARGE \bf%
LLM-as-BT-Planner: Leveraging LLMs for Behavior Tree Generation in Robot Task Planning
}

\author{
Jicong Ao$^{1}$,
Fan Wu$^{1*}$, 
Yansong Wu$^{1}$, 
Abdalla Swikir$^{1,2}$, 
Sami Haddadin$^{1,2}$ 
\thanks{
$^*$ Corresponding author: Fan Wu ({\tt\small f.wu@tum.de}). 

$^{1}$ Munich Institute of Robotics and Machine Intelligence (MIRMI), Technical University of Munich, Germany. 

$^{2}$ Mohamed Bin Zayed University of Artificial Intelligence, Abu Dhabi, UAE.

Project code: \href{https://github.com/ProNeverFake/kios}{https://github.com/ProNeverFake/kios}
}%
}

\renewcommand{\baselinestretch}{0.850}

\let\oldtwocolumn\twocolumn
\renewcommand\twocolumn[1][]{%
    \oldtwocolumn[{#1}{
    \vspace{-20pt}
    \begin{center}
           \includegraphics[width=0.85\textwidth]{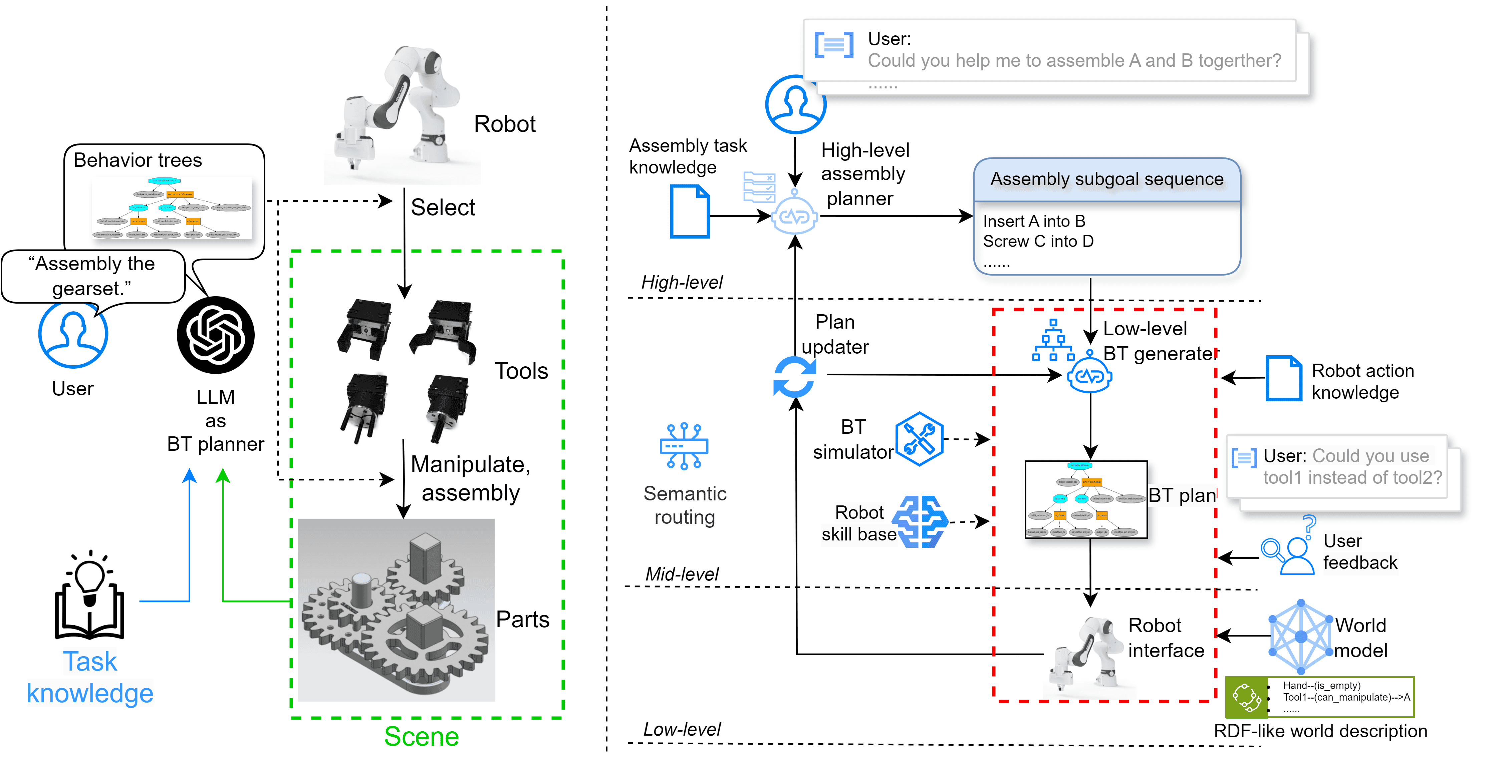}
           \captionof{figure}{\textbf{LLM-as-BT-Planner}. 
           Our framework leverages LLMs to generate executable BTs based on scene states and user instructions directly. \textbf{Left}: The conceptual overview of the framework. The LLM takes user instructions as input, generating BT-based task plans based on the scene information and the task knowledge. Unlike pick-and-place tasks that only require simple robot-object interaction, the assembly tasks require the robot to select appropriate tools to manipulate and assemble the parts. \textbf{Right}: The framework's workflow, including high-level assembly task decomposition, mid-level behavior tree planning, and low-level action execution. The red dashed rectangular indicates the part that our proposed methods can substitute.}       \label{fig:headline}
    \end{center}
    \vspace{-5pt}
    }]
}

\begin{document}

\maketitle

\thispagestyle{empty}
\pagestyle{empty}
\begin{abstract}
Robotic assembly tasks remain an open challenge due to their long horizon nature and complex part relations. Behavior trees (BTs) are increasingly used in robot task planning for their modularity and flexibility, but creating them manually can be effort-intensive. Large language models (LLMs) have recently been applied to robotic task planning for generating action sequences, yet their ability to generate BTs has not been fully investigated. To this end, we propose LLM-as-BT-Planner, a novel framework that leverages LLMs for BT generation in robotic assembly task planning. Four in-context learning methods are introduced to utilize the natural language processing and inference capabilities of LLMs for producing task plans in BT format, reducing manual effort while ensuring robustness and comprehensibility. Additionally, we evaluate the performance of fine-tuned smaller LLMs on the same tasks. Experiments in both simulated and real-world settings demonstrate that our framework enhances LLMs' ability to generate BTs, improving success rate through in-context learning and supervised fine-tuning.
\end{abstract}

\section{INTRODUCTION}
Sequential manipulation planning has been a critical imperative to achieve a higher level of autonomy in robotics. 
Classical approaches to address task planning problems, such as Planning Domain Definition Language (PDDL)~\cite{McDermott1998PDDL}, are based on symbolic formalisms to search for transition paths in the state space to reach task goals. 
In practice, such task plans are often programmed as Finite State Machines, which incorporate expert knowledge specifying control and execution details. 
Due to its limitation in scalability \cite{iovinoSurveyBehaviorTrees2022}, behavior trees (BTs), which represent policies in a state-implicit, hierarchical tree structure, have gained increasing popularity for complex task planning. 
Its advantages of modularity, reusability, and reactivity make it a more desired formalism for long-horizon manipulation tasks. 

Despite being more efficient to program, maintain, and modify than Finite State Machines \cite{iovino2023programming}, manually programming BTs still requires significant effort.
To automate the generation of BT-based task plans in the context of sequential robot manipulation, progress has been made by using \il{\item symbolic planning \cite{colledanchiseBlendedReactivePlanning2019}, \item learning from demonstration \cite{French2019learningBT}, and \item reinforcement learning \cite{Xu2022BTRL}.} However, these methods still largely rely on manual design for the basic structure or available subtrees of BTs. Methods for further automating generation still need to be further explored.

A new approach to address robot task planning has emerged, propelled by the rapid advancement of Large Language Models (LLMs) and Visual-Language Models (VLMs). 
Research progress such as ProgPrompt \cite{singhProgPromptProgramGeneration2023} and PaLME \cite{driessPaLMEEmbodiedMultimodal2023} has come from exploiting the capability of semantic understanding of LLMs (VLMs) and leveraging their reasoning capability which can be improved via in-context learning \cite{wangSelfInstructAligningLanguage2023,weiChainofThoughtPromptingElicits,yaoTreeThoughtsDeliberate2023,ning2024skeletonofthought}.
Few attempts have been made to leverage LLMs for BT generation, where the LLMs are mainly utilized to transfer human instructions into task specifications and initialize a BT expansion algorithm \cite{LLMBT}.
However, a framework that can effectively exploit LLMs to directly generate complex robot task plans represented by BTs has not been seen. 

To this end, we propose \textbf{LLM-as-BT-planner}, an LLM-based BT generation framework to leverage the strengths of both
for sequential manipulation planning, in which different approaches are applied to enhance BT generation performance. To enable human-robot collaborative task planning and enhance intuitive robot programming by nonexperts, the framework takes human instructions to initiate subgoal sequence generation and human feedback to refine BT generation in runtime. 
Moreover, the improvement of LLMs' performance in generating BTs after fine-tuning is investigated based on two different task types.
The presented framework is tested on a real robotic assembly use case, which uses a gearset model from the Siemens Robot Assembly Challenge. We use a single manipulator with a tool-changing mechanism, a common practice in flexible manufacturing, to facilitate robust grasping of a large variety of objects.

To summarize, the main contributions of this work are as follows:
\begin{enumerate}
    \item We established \textbf{a novel framework} to generate fully executable BTs using LLMs in complex robotic assembly tasks with a set of \textbf{performance metrics} for evaluation. 
    The BT generation pipeline includes (i) task decomposition from language instructions to sequence of subgoals, (ii) converting subgoals to parameterized task plans, and (iii) generating BTs given task plans.    
    \item We proposed \textbf{four in-context learning methods} and applied \textbf{supervised fine-tuning} to improve LLMs' performance in BT generation tasks based on our framework, showing its flexibility and expressiveness.    
    \item The performance of the proposed methods is evaluated in a systematic way. Our experiment results demonstrate that \textbf{the human-in-the-loop approach outperforms other in-context learning methods}. While fine-tuning improves the executability of generated BTs, reflecting better in-context understanding and structural generation, it has minimal effect on boosting the success rate of BT generation for smaller (compared to GPT-4) LLMs.  
\end{enumerate}

\section{RELATED WORK}

\mysection{Classical Planning in Robotics:}{
Classical planning relies on symbolic representations, with PDDL \cite{McDermott1998PDDL} being the most widely used domain language. It generates action sequences for small-scale, well-defined problems, leveraging search algorithms to reach the goal state from the initial state under the closed-world assumption 
\cite{helmertFastDownwardPlanning2006, helmertConciseFinitedomainRepresentations2009, wallyProductionPlanningIEC2019}. Researchers have also tried to combine discrete symbolic space planning with continuous geometric space sampling \cite{garrettPDDLStreamIntegratingSymbolic2020, khodeirLearningSearchTask2023}.
However, as shown in \cite{wallyProductionPlanningIEC2019}, large-scale applications of classical planning are still limited due to the exponential increase of complexity in the state space.
Furthermore, the requirements for extensive domain-specific knowledge and the open-loop nature of the generated plan severely reduce its performance in long-horizon task planning.
}


\mysection{Behavior Trees in Robotics:}{The BT framework is a control architecture integrating real-time response with modular system development \cite{ogrenBehaviorTreesRobot2022}. 
Benefiting from its features like the state-free tree structure \cite{iovinoInteractiveDisambiguationBehavior2022}, flexible modification \cite{iovinoSurveyBehaviorTrees2022}, and good interpretability \cite{biggarExpressivenessHierarchyBehavior2021}, it has become a potential task plan representation in the robotics field. 
Since the work of \cite{colledanchiseBlendedReactivePlanning2019}, BTs have been increasingly applied in the robot planning field with respect to condition dependency \cite{liAdaptiveBehaviorTrees2022}, success belief \cite{safronovTaskPlanningBelief2020a}, reactiveness \cite{caiBTExpansionSound2021}, and logical formal check \cite{giunchigliaConditionalBehaviorTrees2019}. Other research focuses on generating valid BTs by applying genetic programming 
\cite{styrudCombiningPlanningLearning2022, colledanchiseLearningBehaviorTrees2019}, grammatical programming \cite{dengLearningBehaviorTrees2023, scheideBehaviorTreeLearning2021}, and value function-based policies \cite{zhangCombiningBehaviorTrees2017}.
However, the application of BT still highly depends on pre-definition and pre-programming.
Furthermore, BTs can only react to scenarios that are explicitly defined in their structure. To generalize their application, an effective feedback-adjustment mechanism is necessary.
}

\mysection{LLM-based Robot Task Planning:}{
Many studies have been conducted to integrate LLMs with robot task planning. Some research tries to utilize LLMs in generating complete problem descriptions \cite{zhouISRLLMIterativeSelfRefined2023, liuLLMEmpoweringLarge2023, guanLeveragingPretrainedLarge2023, LLMBT, silverGeneralizedPlanningPDDL2023a} and tracking world states \cite{chenLLMStateExpandableState2023, yonedaStatlerStateMaintainingLanguage2023, singhProgPromptProgramGeneration2023} to support the planning process.
Others leverage LLMs in grounding language instruction for task understanding \cite{driessPaLMEEmbodiedMultimodal2023}, policy selection \cite{ahnCanNotSay2022}, and control command generation \cite{zitkovich2023rt}.
However, the task plans generated by the aforementioned methods are mainly action sequences. A framework does not yet exist to explore leveraging LLMs to directly generate task plans represented by BTs. The ability of LLMs to generate BTs for robot assembly tasks has also not been comprehensively investigated.
}

\section{LLM-as-BT-PLANNER} \label{sec:framework}
\subsection{Problem Statement}

The task planning problem is formulated as the tuple \((\mathcal{O}, P, C, R, A, 
T, 
I, G, t)\). The set \(\mathcal{O}\) comprises all objects available in the environment. Properties of these objects, denoted by \(P\), inform object affordances and availability. Constraints between objects in \(\mathcal{O}\) are represented by \(C\), while relationships between objects are denoted by \(R\). The set of executable actions, \(A\), can be applied to change the environment state, which is defined as \(s \in S\). A state \(s\) is a specific assignment of all object properties, constraints, and relationships, with \(S\) being the set of all possible assignments. Constraints in \(C\) remain unchanged by actions in \(A\), whereas the actions can alter relationships in \(R\). 
The transition model is represented by \(T: S \times A \rightarrow S\).
The initial and goal states are denoted by \(I\) and \(G\). Instead of the specific goal state \(g \in G\), the agent receives a corresponding high-level task description \(t\) in natural language. 
With \(\mathcal{O}\), \(P\), \(C\), \(R\), \(A\), and \(I\) known, Our task is to generate a BT that brings the world state from \(I\) to \(G\) by its execution.

\subsection{Framework Design and Workflow}

Our framework design is
shown in Fig.~\ref{fig:headline}. In the high-level part, user instructions, as inputs, are processed by a high-level LLM-based assembly planner at the beginning for 
task decomposition, generating a sequence of subgoals.
In this step, the necessary task knowledge is provided in natural language similar to the way in \cite{wangPlanandSolvePromptingImproving2023}. 
In the mid-level part, the subgoal sequence from the high-level part is taken by an LLM-based BT generator as planning targets. 
Notably, we adopt the BT definitions and the subtree structure in \cite{colledanchiseBlendedReactivePlanning2019}, which are effective for constructing asynchronous BTs.
During the BT generation process, the knowledge of robot actions and world predicates is utilized, which is written in a PDDL-like form with natural language explanations.
The world state is also an important reference for BT generation, which is represented in an RDF-like format as in \cite{singhProgPromptProgramGeneration2023, yonedaStatlerStateMaintainingLanguage2023}. 
In the low-level part, the robot interface loads the generated BT 
and executes it with the help of the world model, which provides the world state and the spatial information of the objects in the environment. 

Between BT generation and execution, it is possible to involve human feedback for runtime BT replanning.
To mitigate execution risks, simulation feedback can help pre-adjust BT plans before their execution. 
A skill library equips the executor with diverse robotic actions, ensuring the accuracy and adaptability of assembly task execution. Semantic routers are applied to guide the workflow by calculating the embedding distance between user inputs and the corresponding exemplary input samples.

\section{IN-CONTEXT LEARNING AND SUPERVISED FINE-TUNING} \label{sec:methods}
\subsection{In-context Learning Methods}

Four LLM-based BT generation methods are designed to be applied in our framework, as introduced in Scheme 1 - 4 below. 

\begin{figure}[h]
  \centering
  \vspace{-15pt}
  \begin{subfigure}[b]{0.99\linewidth}
    \centering
    \begin{subfigure}[b]{0.4\linewidth}
      \includegraphics[width=\linewidth]{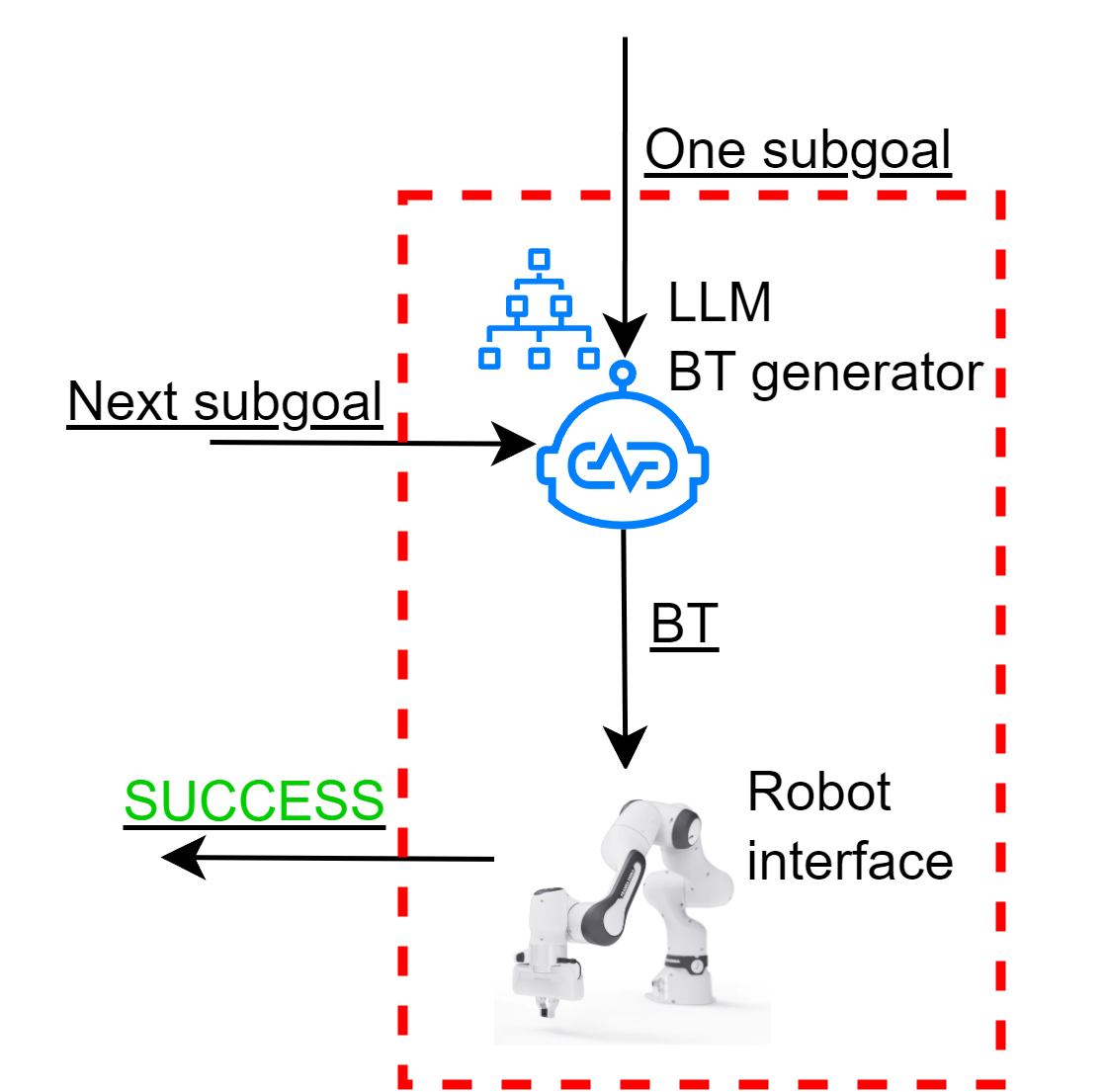}
      \caption{One-step generation}\label{subfig:onestep}
    \end{subfigure}%
    \hfill 
    \begin{subfigure}[b]{0.6\linewidth}
      \includegraphics[width=\linewidth]{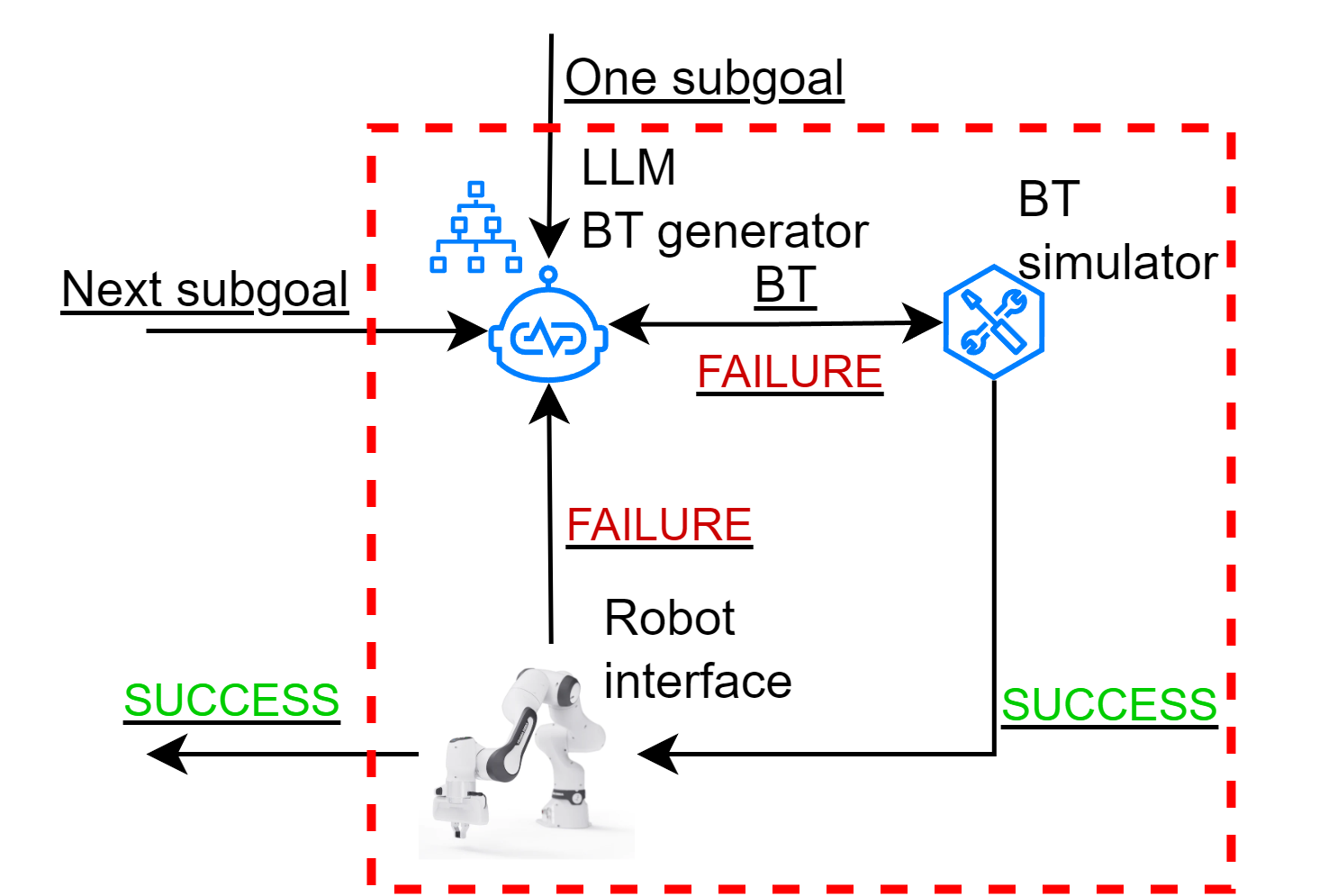}
      \caption{Iterative generation}\label{subfig:iter}
    \end{subfigure}
  \end{subfigure}\\[2ex]
  
  \begin{subfigure}[b]{0.99\linewidth}
    \centering
    
    \begin{subfigure}[b]{0.55\linewidth}
      \includegraphics[width=\linewidth]{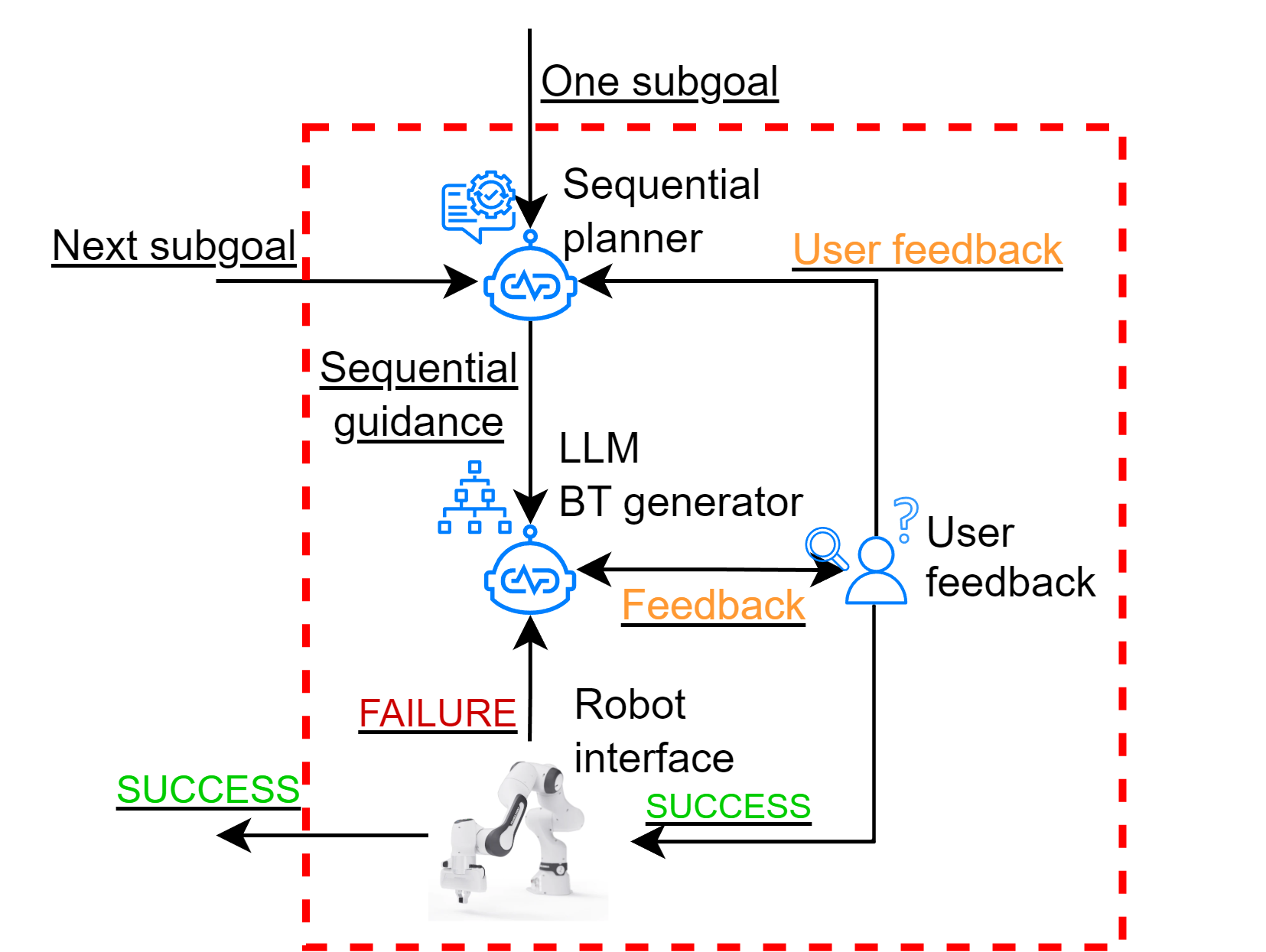}
      \caption{Human-in-the-loop generation}\label{subfig:hil}
    \end{subfigure}%
    \hfill 
    \begin{subfigure}[b]{0.45\linewidth}
      \includegraphics[width=\linewidth]{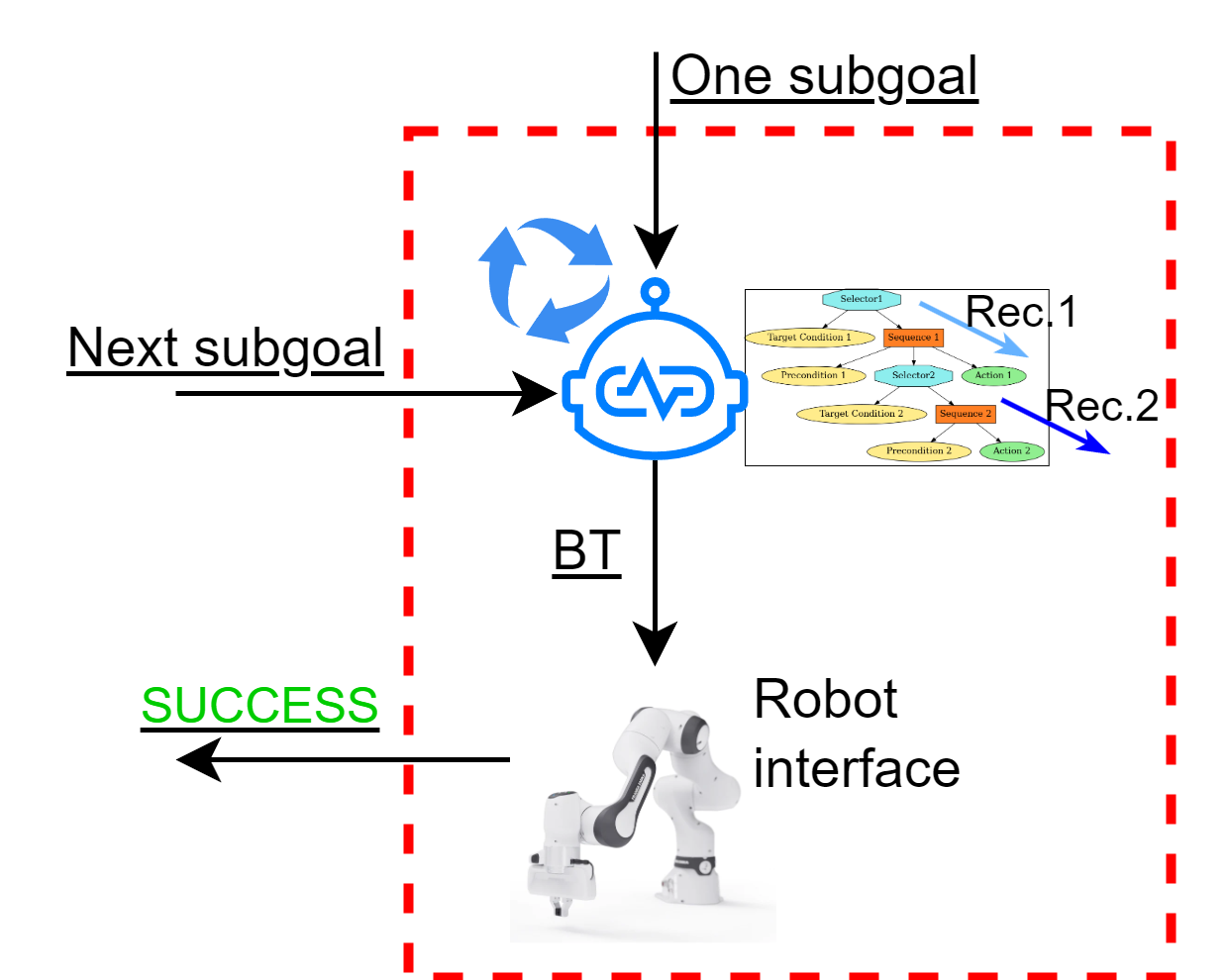}
      \caption{Recursive generation}\label{subfig:rec}
    \end{subfigure}
  \end{subfigure}\\[1ex]

  \caption{\textbf{The four proposed in-context learning methods}, where the red dashed rectangles show the place in the workflow (shown in Fig.~\ref{fig:headline}) that the contents from different methods can substitute.} \label{fig:four methods}
  \vspace{-14pt}
\end{figure}

\subsubsection*{\schemeref{onestep} - One-step generation (Fig.~\ref{subfig:onestep})}\manuallabel{scheme:onestep}{1}
As shown in the figure, this method uses an LLM-based BT generator to \textbf{generate BTs directly for an assembly subgoal coming from the upstream module}. 
Given the initial state, the subgoal, and the knowledge about actions and BTs, the LLM generates an entire BT that can achieve the target state from the initial state. 
The generated BT will be passed directly to the robot executor for execution, and the result will be passed back to the upstream plan updater.
Should the BT fail in execution, a \textit{FAILURE} signal will be passed to prevent the plan update, leading to a replan for the last assembly step.
This method is proposed as a basic LLM-based BT generation method, which is then improved by the methods introduced below.
    
\subsubsection*{\schemeref{iterative} - Iterative generation (Fig.~\ref{subfig:iter})}\manuallabel{scheme:iterative}{2}
This method \textbf{leverages the result from the BT simulation} to help rectify and regenerate the BT. 
The process of generating the entire BT is the same as that in Scheme 1.
The generated BT is then passed to the simulator for execution, which returns an execution result based on the assumption that all the action nodes finally return a \textit{SUCCESS} signal. 
With a \textit{FAILURE} result, which is mainly due to structural error or logic inconsistency, the simulator will provide predefined failure reasons, which are utilized by the generator to generate a new BT.
With \textbf{simulation feedback} introduced, this method leverages the inference ability of LLMs to improve the BTs based on simulation feedback.

\subsubsection*{\schemeref{human} - Human-in-the-loop generation (Fig.~\ref{subfig:hil})}\manuallabel{scheme:human}{3}

The human-in-the-loop generation \textbf{adopts a human feedback step} 
to provide precise suggestions for BT rectification and improvement. 
This method applies a sequential planner to generate a bullet plan for the upstream action step with natural language explanations first (inspired by \cite{weiChainofThoughtPromptingElicits}), which is then passed to the BT generation module to help guide the generation of the BT. 
After BT generation, a user feedback step is introduced to provide feedback for improving the generated BT plan. The same step is also introduced after the execution of the BT, which allows the user to manipulate the planning and execution process flexibly. 
By \textbf{introducing user feedback}, this method aims to 
provide precise feedback on the generation result
and allows for BT modification based on natural language feedback in the planning phase.

\subsubsection*{\schemeref{recursive} - Recursive generation (Fig.~\ref{subfig:rec})}\manuallabel{scheme:recursive}{4}

The recursive generation method \textbf{utilizes an algorithm to guide the BT generation process},
dividing the full generation process into three different LLM invoking steps, i.e., {\tt MakePlan}, {\tt MakeTree}, and {\tt PredictState}, and generating the complete BT by recursively invoking them.
In the {\tt MakePlan} step, the LLM generates an action sequence to satisfy the current unfulfilled condition. The {\tt MakeTree} step generates a subtree for the first action in that action sequence and replaces the unfulfilled condition node with it. The {\tt PredictState} step pushes forward the world state by predicting the world state after the execution of the new subtree, updating the world state for following planning.
The detail of the expansion process is shown in Algo.~\ref{algo}.
By \textbf{dividing the process and applying algorithm guidance}, this method aims to improve the accuracy and robustness of BT generation with higher resource consumption.

\begin{algorithm}[!htb]
\scriptsize 
\caption{Behavior Tree Expansion Algorithm} \label{algo}
\begin{algorithmic}[1]

\Function{ExpandBehaviorTree}{$node\_list, s\_0$}
    \State $s'\_0 \gets s\_0$
    \For{each $node_i$ in $node\_list$}
        \State $g_i \gets$ \Call{GetGoal}{$node_i$}
        \State $plan_i \gets$ \Call{MakePlan}{$s'\_{i-1}, g_i$}
        \If{$\text{len}(plan_i) > 0$}
            \State $s'\_i \gets$ \Call{PredictState}{$s'\_{i-1}, plan_i$}
            \State $a_i \gets plan_i[-1]$
            \State $tree_i \gets$ \Call{MakeTree}{$a_i$}
            \State $new\_node\_list \gets$ \Call{GetCondChildren}{$tree_i$}
            \State \Call{ExpandBehaviorTree}{$new\_node\_list, s'\_i$}
        \Else
            \State $s'\_i \gets s'\_{i-1}$
        \EndIf
    \EndFor
\EndFunction

\end{algorithmic}
\end{algorithm}

\subsection{Behavior Tree Generation Tasks for Fine-tuned LLMs} \label{sft}

Two BT generation task types are designed to investigate the performance enhancement of LLM fine-tuning in different perspectives of BT generation tasks.

\subsubsection*{Unit-tree Generation Tasks (Fig.~\ref{subfig:utg})}
The unit-tree generation task is a subtask from the recursive generation method. 
Given an action, full action definitions and BT structure requirements, this task asks the LLM to \textbf{interpret the action into a unit BT with its corresponding preconditions and action node}. 
This task aims to \textbf{evaluate the structural output capability and the context learning ability of LLMs in generating BTs}.

\begin{figure}[h]
  \centering
  \vspace{-5pt}
  \begin{subfigure}[b]{0.99\linewidth}
    \centering
    
    \begin{subfigure}[b]{\linewidth}
      \includegraphics[width=\linewidth]{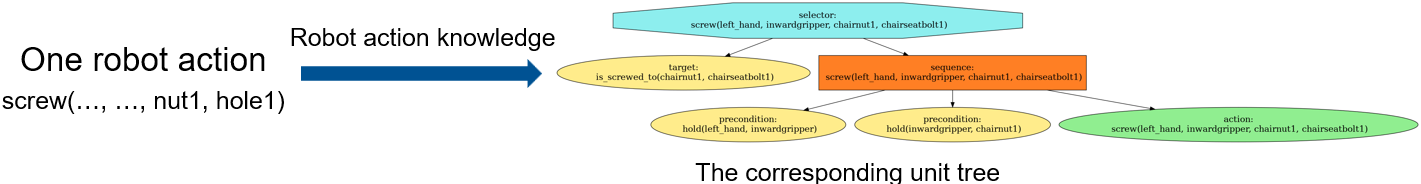}
      \caption{An example of unit-tree generation tasks}\label{subfig:utg}
    \end{subfigure}%
    \hfill 
    \begin{subfigure}[b]{\linewidth}
      \includegraphics[width=\linewidth]{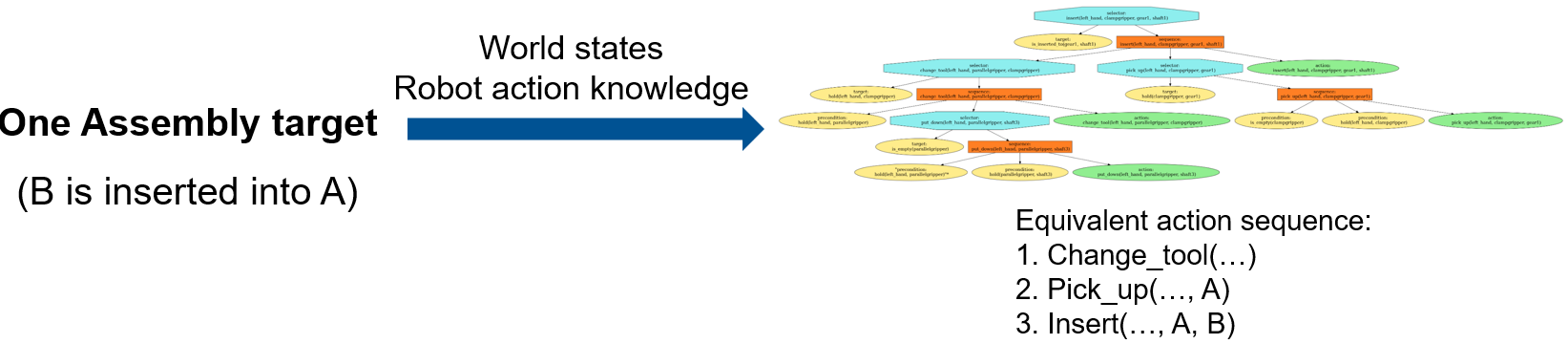}
      \caption{An example of one-step generation tasks}\label{subfig:osg}
    \end{subfigure}
  \end{subfigure}\\[1ex]

  \caption{The examples of (a) the unit-tree generation tasks and (b) the one-step generation tasks. The BTs in the figures are only for illustrative purposes to visualize the processes and their complexity. If the readers are interested in the details of the BTs in the examples, we refer to high-resolution images in our code repository.} \label{fig:four methods}
  \vspace{-17pt}
\end{figure}

\subsubsection*{One-step Generation Tasks (Fig.~\ref{subfig:utg})}
The One-step generation task is the task from the one-step generation method. 
Given the initial state, the goal description, and the necessary knowledge, this task requires the LLMs to \textbf{generate an entire BT-represented task plan that can achieve the goal state after its execution}. 
Besides the structural output ability and the context learning ability of LLMs, this task also \textbf{evaluates the inference ability of LLMs}, requiring LLMs to handle the dependency and nesting relations between sub-BT-represented actions.

\section{EXPERIMENTS AND RESULTS} \label{sec:results}

\subsection{Experiment Setup}

\begin{figure}[htb]
\centering
\vspace{-10pt}
\includegraphics[width=\linewidth]{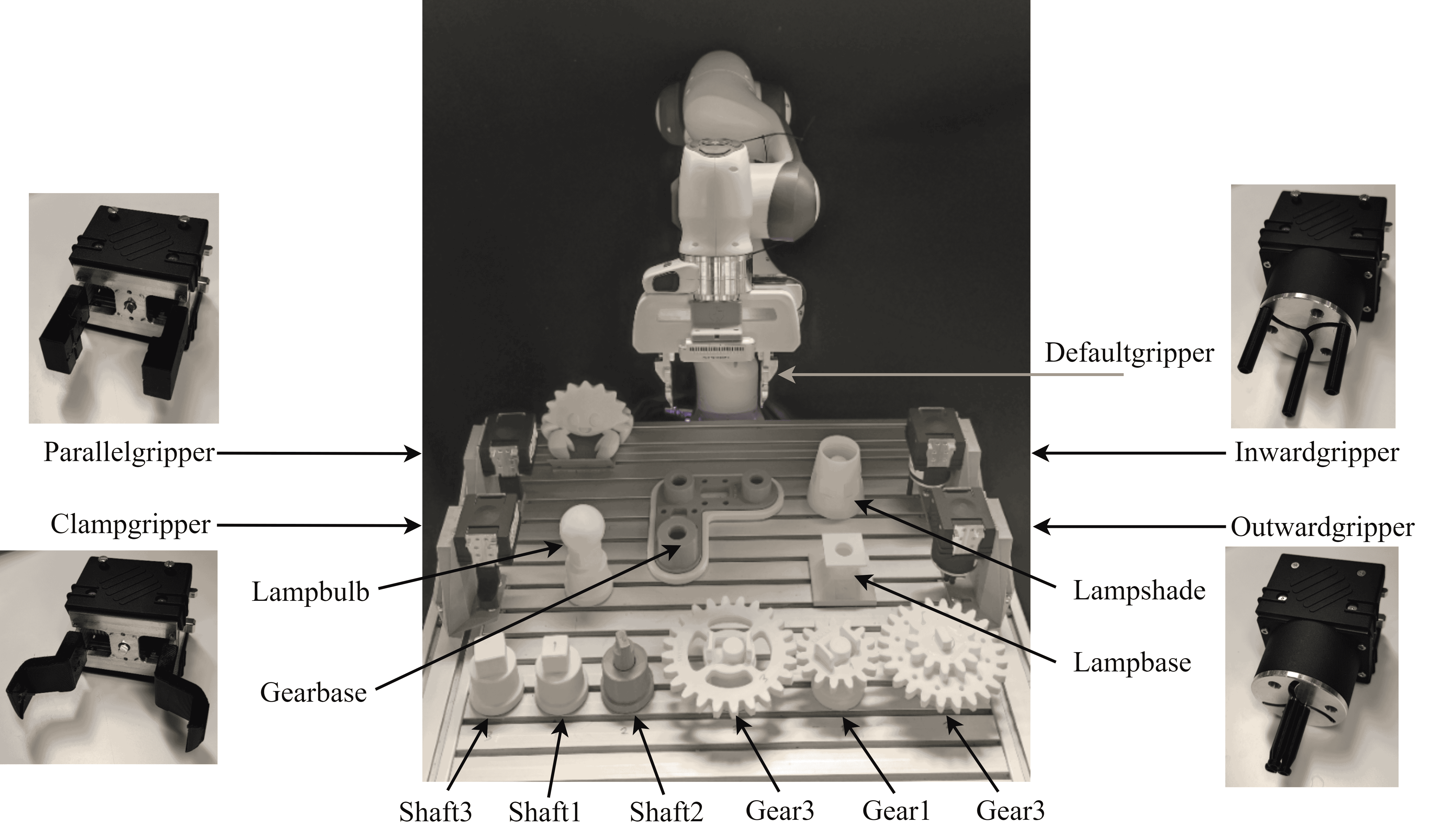}
\caption{\textbf{Experiment setup} 
with a franka panda robot, four tool cubes from Leverage, and a gearset from the Siemens Robot Assembly Challenge. 
} \label{fig:usecase_details}
\vspace{-10pt}
\end{figure}

The experiment setup of physical robot validation is shown in Fig.~\ref{fig:usecase_details}. Operation space control is applied to a single 7-DoF Franka Emika Panda robot arm via Franka Control Interface (FCI) \cite{10610835}. The computed torques come from a Cartesian adaptive force-impedance controller.
To execute the actions in BTs, the action context is parsed to another custom control software, i.e., the skill base, to map the action to a pre-defined skill. 
PyTrees and WebSocket are used to implement asynchronous BTs. 
The world model integrates a self-implemented relation graph with Neo4j for world state management and visualization. 


The details of the use case are shown in Fig.~\ref{fig:usecase_details}.
In the experiment, the robot is provided with tool cubes classified as {\tt parallelgripper}, {\tt clampgripper}, {\tt outwardgripper}, and {\tt inwardgripper} according to their self-designed fingertips. 
The generation methods are evaluated using the Siemens Robotic Assembly Challenge gearset use case, involving a gear base, three shafts, and three gears. 
The task is to assemble the shafts into the gearbase and insert the gears into their corresponding shafts to mesh. 
Here, the second shaft fixed on the gearbase is modified to be screwable to diverse the necessary robot skill types. 
Different LLMs are employed for BT generation across all four proposed approaches. 
For validation, the human-in-the-loop generation method is adopted for its superior performance and the experiment convenience its human feedback brings.

For the states in our experiments, the property set \(P\) includes properties like {\tt isEmpty}. 
The constraint set \(C\) includes {\tt canManipulate} and {\tt isInsertable}. 
The relation set \(R\) includes {\tt isInsertedTo} and {\tt Hold}. 
The object set \(\mathcal{O}\) includes {\tt left hand}, {\tt shaft1}, {\tt shaft2}, {\tt shaft3}, {\tt gearbase hole1}, {\tt gearbase hole2}, {\tt gearbase hole3}, {\tt gear1}, {\tt gear2}, {\tt gear3}, and tools such as {\tt defaultgripper}, {\tt clampgripper}, and {\tt inwardgripper}. 
The action set \(A\) includes {\tt insert}, {\tt screw}, {\tt place}, {\tt put down}, {\tt change tool}, and {\tt pick up}. 
Experiments are also conducted on the chair and lamp use cases from the Furniture Assembly Benchmark in \cite{heoFurnitureBenchReproducibleRealWorld2023}, which ask for assembling a chair and a lamp with objects like {\tt lamp base}, {\tt lamp bulb}, {\tt chair leg}, and {\tt chair seat}.

\subsection{Fine-tuning Process} \label{sec:ftprocess}

Two open-source LLMs, \textbf{Mistral-7B} and \textbf{Llama2-13B-chat}, are selected and fine-tuned for the two task types. A \textbf{GPT-3.5} model is also fine-tuned to show the performance of LLMs with larger parameter amounts. 
For the unit-tree generation task, training data are collected from the roll-out records of the recursive generation method with Siemens' gearset assembly task as the use case. 
For the one-step generation task, the data collected from the one-step generation method are adopted as training data.
In the training phase, both Mistral-7B and Llama2-13B-chat are trained using the Llama-factory framework \cite{zheng2024llamafactory}. For unit-tree generation tasks, Mistral-7B and Llama2-13B-chat are trained for 10 epochs with a learning rate of $1 \times 10^{-4}$, while for One-step generation tasks, they are trained for 15 epochs with a learning rate of $5 \times 10^{-5}$. GPT-3.5 is fine-tuned with a learning rate multiplier of 0.05 for 3 epochs for unit-tree generation tasks, while for one-step generation tasks, the same learning rate multiplier is applied for 10 epochs.
After training, the LLMs' performance on both task types is validated with the data collected from the chair and lamp use cases of the furniture assembly benchmark \cite{heoFurnitureBenchReproducibleRealWorld2023}. 

\subsection{Evaluation Metrics} \label{sec:metrics}


\begin{description}
    \item[SR] Success Rate. 
    A generation can be taken as a success only if the generated BT is executable, logically coherent, and can achieve the goal state. 
    This metric shows the overall capability of the method to generate a correct BT.
    \item[LC] Logical Coherence. This means the execution order inside the BT aligns with its equivalent action sequence without precondition violation.
    This metric shows the inference capability of the LLM in the corresponding generation method.
    \item[Exec] Executability. 
    This means the BT follows the regulated format and can be executed. 
    This metric shows the LLMs' capability of generating structured outputs using the corresponding methods.
    \item[GD] Generation Duration for generating an entire BT. This metric shows the time consumption of the method.
    \item[TC] Token Consumption for generating an entire BT. This metric shows the resource consumption of the method. 
\end{description}

\subsection{In-context Learning Results} \label{sec:in-context result}

The evaluation of four proposed methods, as detailed in Table \ref{tab:2}, highlights their varying efficiency and effectiveness in generating BTs within the framework.

\textbf{GPT-4 shows good in-context learning and structured output generation capabilities in generating BTs.} 
Across all four methods, GPT-4 generates BTs that are highly executable (13/17 for the recursive method and 17/17 for the other three methods). For logical coherency and success rate, the results show that GPT-4 can tackle over half of the total 17 tasks (the lowest is 12/17 for both the iterative method and the one-step method.).
Specifically, the one-step generation method exhibits perfect BT executability and a success rate score of 12 out of 17. 
By looking into the failure cases, most failures are due to insufficient tree depth and the lack of well-defined actions, which shows the limitation of this method. 

\begin{table}[H]
\vspace{-3pt}
\scriptsize 

\begin{threeparttable}
\caption{Result comparison of the four in-context learning methods.
}
\label{tab:2}
\setlength\tabcolsep{1.5pt} 

\begin{tabular*}{\columnwidth}{@{\extracolsep{\fill}} l ccccc}
\toprule
     Method &  
     \multicolumn{3}{c}{Accuracy} & GD(sec.) & TC \\
\cmidrule{2-4}
     & SR & LC & Exec \\
\midrule
     One-step & 12/17 & 12/17 & 17/17 & 49.11 & 5074.96 \\
     Iterative & 12/17 & 12/17 & 17/17 & 48.52 & 7770.13\\
     Human-in-the-loop & 16/17 & 16/17 & 17/17 & 85.02 & 7483.34\\
     Recursive & 13/17 & 17/17 & 13/17 & 231.04 & 50229.96\\
\bottomrule
\end{tabular*}

\end{threeparttable}
\vspace{-8pt}
\end{table}

\textbf{Non-specific simulation feedback doesn't contribute to performance improvement.}
The iterative method does not show an advantage over one-step generation because all BTs generated by both methods in the test cases are executable, which presents the ineffectiveness of simulation feedback. This means the predefined non-specific failure reasons in the feedback can not be leveraged by the iterative method.

\textbf{Specific and directed user feedback significantly optimizes LLMs' performance.}
The human-in-the-loop method demonstrates a significant improvement in logical coherence (16/17 compared to 12/17 of the one-step method) and executability (17/17, better than 13/17 of the recursive method) because of the incorporation of precise user feedback, which also leads to larger average generation duration (85.02 seconds) and token consumption (7483.34 tokens). 
Notably, this method also allows the introduction of new natural language-represented knowledge for BT generation.
The LLM leverages its powerful natural language processing capabilities to prioritize new input, such as changes in tool-object compatibility, and generates BTs based on this information rather than relying on outdated knowledge. This allows for efficient replanning without altering the knowledge base.

\textbf{Algorithm guiding enhances BT generation performance with unstable structured output performance and more resource consumption.}
The recursive method, while ensuring high logical coherence (17/17, the highest among all methods), incurs the largest resource consumption (231.04 seconds for generating duration and 50229.96 for token consumption), reflecting its thoroughness in distributing the generation task across multiple recursive LLM invokes. 
These results highlight this method's trade-offs between generation time, complexity, and accuracy.
Furthermore, it also shows the worst executability result (13/17). This is caused by the recursive invocations in the generation process, which magnifies the instability of LLMs' generation capability.

To summarize, \textbf{the human-in-the-loop approach} stands out for its high success rate and balance against efficiency and token consumption. 
The recursive method, though consuming a huge amount of time and tokens, shows an excellent ability to generate logically coherent BTs, which may be more beneficial when using smaller, fine-tuned, locally deployed LLMs like Llama-2-13B instead of GPT-4.

\subsection{Fine-tuning Results}

\begin{figure*}[ht]
\centering
\vspace{-10pt}
\includegraphics[width=0.9\linewidth]{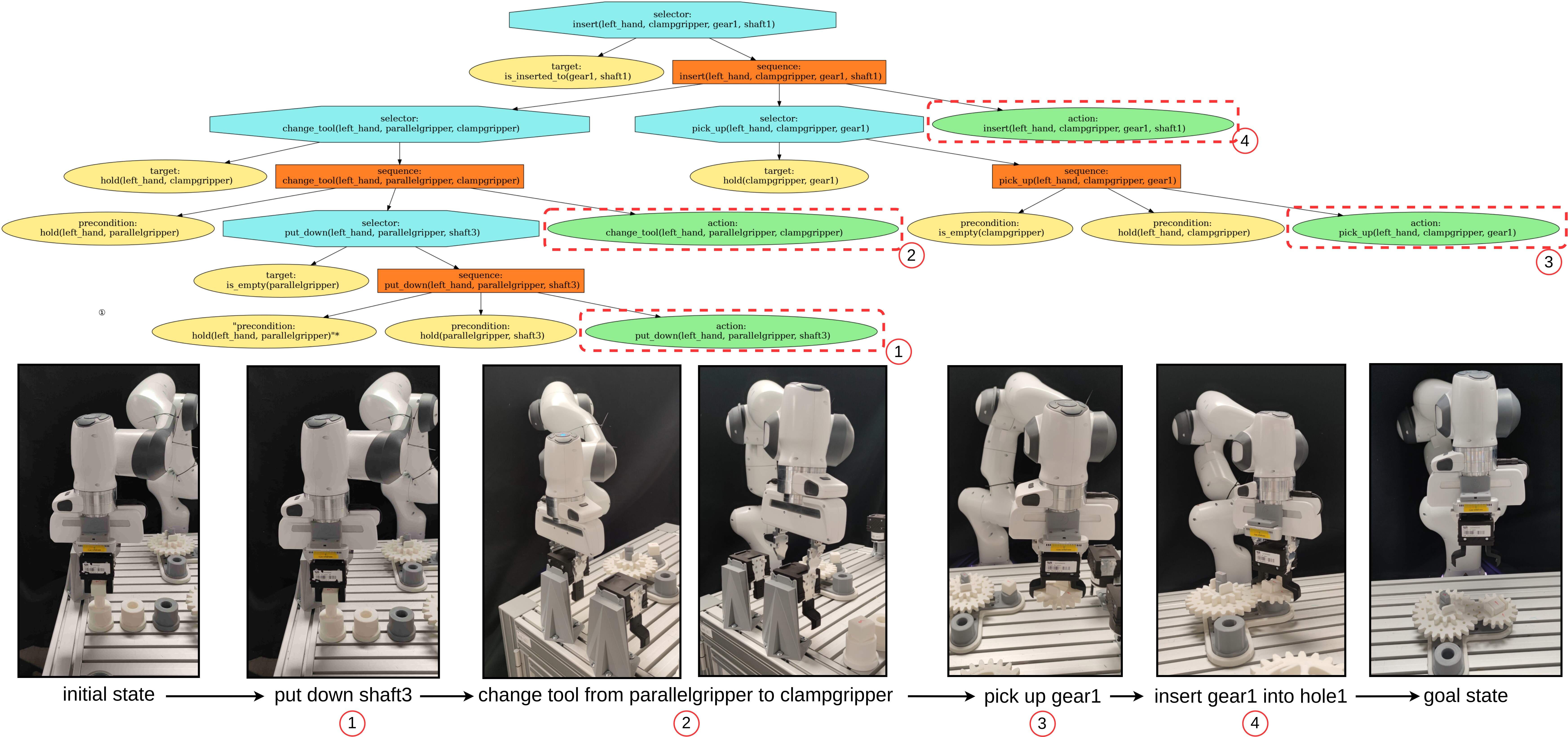}
\caption{\textbf{Robotic assembly of a gear set}. The generated BT and the corresponding sequence of actions are shown. The order of actions is labeled by number and shown from left to right, while their corresponding action nodes in the BT are colored green.} \label{fig:exe}
\vspace{-17pt}
\end{figure*}

The performance of pre-trained and fine-tined LLMs in both task types is shown in Table~\ref{tab:ftone}. 
The success rate of the unit tree generation task is not evaluated because the unit tree generation tasks are not state-related and can not be evaluated alone.

\begin{table}[h]
\vspace{-3pt}
\scriptsize 
\begin{threeparttable}
\caption{Comparison of BT generation results of pre-trained and fine-tuned LLMs in both task types.
}
\label{tab:ftone}
\setlength\tabcolsep{0pt} 

\begin{tabular*}{\columnwidth}{@{\extracolsep{\fill}} l c c ccccc}
\toprule
     Task Type & Model & Fine-tuned & 
     \multicolumn{3}{c}{Accuracy} & GD(sec.) & TC \\
\cmidrule{4-6}
     & & &SR & LC & Exec \\
\midrule
    Unit tree& GPT-4 & No & - & 10/10 & 10/10 & 14.45 & 2229.00 \\
     Unit tree& GPT-3.5 & No & - & 10/10 & 10/10 & 6.07 & 2220.20 \\
     Unit tree& GPT-3.5 & Yes & - & 10/10 & 10/10 & 6.17 & 2214.00 \\
     Unit tree& Mistral-7B & No & - & 3/10 & 7/10 & 14.64 & 1993.00 \\ 
     Unit tree & Mistral-7B & Yes & - & 9/10 & 10/10 & 14.35 & 1920.30 \\
     Unit tree & Llama-13B-chat & No & - & 5/10 & 6/10 & 16.30 & 2142.80 \\ 
     Unit tree & Llama-13B-chat & Yes & - & 9/10 & 10/10 & 17.56 & 1964.60 \\
     \cmidrule{1-8}
     One-step & GPT-4 & No & 9/10 & 10/10 & 10/10 & 45.48 & 4515.00 \\
     One-step & GPT-3.5 & No & 1/10 & 1/10 & 2/10 & 11.30 & 4352.00 \\
     One-step & GPT-3.5 & Yes & 1/10 & 1/10 & 9/10 & 10.79 & 4184.00 \\ 
     One-step & Mistral-7B & No & 0/10 & 0/10 & 0/10 & 25.54 & 4139.40 \\
     One-step & Mistral-7B & Yes & 0/10 & 0/10 & 8/10 & 24.60 & 4157.80 \\
     One-step & Llama-13B-chat & No & 0/10 & 0/10 & 5/10 & 26.45 & 4227.30 \\
     One-step & Llama-13B-chat & Yes & 1/10 & 1/10 & 9/10 & 25.72 & 4168.30 \\
\bottomrule
\end{tabular*}

\end{threeparttable}
\vspace{-7pt}
\end{table}

\textbf{Pre-trained large models outperform small models in complex tasks.}
As the table shows, GPT-4 excels across all metrics in both task types, demonstrating its superior ability in in-context learning and structural generation, largely attributed to its extensive parameter scale. In contrast, pre-trained small models, i.e., GPT-3.5, Llama-13B-chat, and Mistral-7B, though showing perfectly in unit tree generation tasks, perform poorly in one-step generation tasks. 
Specifically, vanilla GPT-3.5 gets worse results in logical coherency (1/10) and executability (2/10) compared to GPT-4 (both 10/10), and vanilla Llama-13B-chat performs better than vanilla Mistral-7B in executability (5/10 compared to 0/10). This aligns with the results of the unit tree generation tasks, showing the advantages of pre-trained models with larger parameter amounts in the BT generation tasks of robotic assembly.

\textbf{Fine-tuning significantly improves models’ performance in structured output generation and in-context learning.}
In unit tree generation tasks, Llama2-13B-chat and Mistral-7B both show a mostly perfect performance in both logical coherence (both 9/10) and executability (both 10/10) compared to their vanilla models (5/10 and 3/10 for logic coherency and 6/10 and 7/10 for executability, respectively). 
While in one-step generation tasks, both small models show improvement in executability after being fine-tuned (8/10 and 9/10 compared to 0/10 and 5/10, respectively).
This indicates that the models' capabilities of structural generation and context understanding are enhanced through effective fine-tuning.

\textbf{Even after fine-tuning, small LLMs perform poorly in complex BT generation tasks that require inference.}
For unit tree generation tasks, which primarily assess the in-context learning capability of LLMs, both Llama2-13B-chat and Mistral-7B outperform their corresponding vanilla models in terms of logic coherency. 
For one-step generation tasks that focus on generating BTs with correct internal dependency relations, Llama2-13B-chat, Mistral-7B, and GPT-3.5 show no performance improvement after fine-tuning, achieving success rates of 0/10 or 1/10 for logic coherency. This indicates the limitations of fine-tuning in enhancing the inference capabilities of small LLMs for complex BT generation tasks. 
Given the nearly perfect performance of the larger parameter GPT-4 across all accuracy metrics without fine-tuning, the limited improvement could be attributed to the smaller parameter amount of these models and the insufficiency in training data for one-step generation tasks. 

\subsection{Real Robot Validation}

The real robot validation process is shown in Fig.~\ref{fig:exe}. The BT can be generated by any of the four proposed methods to satisfy the upstream subgoal {\tt insert gear1 into shaft1} and represents its equivalent action sequence: 
\begin{enumerate}
    \item {\tt\small put\_down(left\_hand, parallelgripper, shaft3)},
    \item {\tt\small change\_tool(left\_hand, parallelgripper, clampgripper)},
    \item {\tt\small pick\_up(left\_hand, clampgripper, gear1)},
    \item {\tt\small insert(left\_hand, clampgripper, gear1, shaft1)}.
\end{enumerate}

At the initial state, the {\tt left\_hand} holds the {\tt parallelgripper}, which is holding {\tt shaft3}. With the preconditions satisfied, the action (1) is executed first to satisfy the condition {\tt is\_empty(parallelgripper)}. This allows the action (2) to proceed, which switches the tool 
from {\tt parallelgripper}
to {\tt clampgripper}. The condition {\tt hold(left\_hand, clampgripper)} is fulfilled by this and the action (3) starts to be executed. After this, the preconditions for the action (4), namely {\tt hold(left\_hand, clampgripper)} and {\tt hold(clampgripper, gear1)}, are satisfied, allowing its execution to complete the planning target {\tt is\_inserted\_to(gear1, shaft1)}. In the end, the BT returns \textit{SUCCESS}, indicating that the task {\tt insert gear1 into shaft1} is successfully completed.

The validation experiment demonstrates that the proposed LLM-based BT generation methods enable the framework to generate and execute BT-based robotic assembly plans effectively and efficiently.

\subsection{Limitations}
We summarize the limitations of our work as follows. 
First, the recursive generation method is resource-intensive and needs further improvement.
Second, fine-tuning has shown limited effectiveness in enhancing the inference capabilities of few-parameter LLMs for generating BTs. Possible reasons are the limited amount of model parameters and the insufficient training dataset. 
Furthermore, we note the challenge of using BTs to represent task plans in a large number of steps. As mentioned in \cite{LLMBT} and \cite{colledanchiseBlendedReactivePlanning2019}, conflicts between nodes can happen when BTs have a series of deeply nested steps. 
Although our approach of employing high-level task decomposition has mitigated this issue to a certain extent by reducing the number of steps included in each subgoal, further exploration is required for the broader applicability of our framework.

\section{CONCLUSIONS}
\label{sec:conclusion}

This work introduces LLM-as-BT-planner, an innovative framework that leverages LLMs for BT generation in robotic task planning. Various BT generation methods are explored based on different LLMs using in-context learning and fine-tuning techniques. Experimental evaluations show that our framework effectively generates BTs for robotic assembly tasks, offering a promising solution for complex task planning scenarios. 
Among the four BT generation methods, the human-in-the-loop approach outperforms the others. Fine-tuning improves LLMs' performance in context understanding and structural generation but has an insignificant impact on few-parameter LLMs inference capability. 
This may be due to insufficient training data and limitations in model parameters, which need to be further investigated in future work.

\section*{ACKNOWLEDGMENT}
The authors acknowledge the financial support by the Bavarian State Ministry for Economic Affairs, Regional Development and Energy (StMWi) for the Lighthouse Initiative KI.FABRIK (Phase 1: Infrastructure as well as the research and development program under, grant no. DIK0249). 
The authors acknowledge the financial support by the Federal Ministry of Education and Research of Germany (BMBF) in the programme of "Souverän. Digital. Vernetzt." Joint project 6G-life, project identification number 16KISK002.



\bibliographystyle{myIEEEtran}
\bibliography{IEEEabrv,mybib2022}

\label{last-page}
\end{document}